\documentclass[10pt,twocolumn,letterpaper]{article}

\usepackage[pagenumbers]{cvpr} 

\definecolor{cvprblue}{rgb}{0.21,0.49,0.74}
\usepackage[pagebackref,breaklinks,colorlinks,allcolors=cvprblue]{hyperref}

\usepackage{booktabs}
\usepackage{multirow}
\usepackage{color, colortbl}
\usepackage{subcaption}
\usepackage{tabu}
\usepackage{pifont}
\usepackage{makecell}
\usepackage{paralist}
\usepackage{threeparttable}
\usepackage{enumitem}
\usepackage{hhline}
\usepackage{scalerel,xparse}
\usepackage{amssymb}
\usepackage{amsmath}

\def\paperID{}
\def\confName{CVPR}
\def\confYear{2026}

\definecolor{asparagus}{RGB}{0.53, 0.66, 0.42}
\definecolor{Gray}{gray}{0.5}
\definecolor{LGray}{gray}{0.9}
\definecolor{darkblue}{RGB}{94,110,186}
\definecolor{darkGreen}{RGB}{92, 148, 110}
\definecolor{myblue}{RGB}{14, 121, 178}

\usepackage{etoolbox}
\usepackage{algorithm}
\usepackage{listings}

\usepackage{etoolbox}
\makeatletter
\renewcommand\fs@ruled{%
  \def\@fs@cfont{\bfseries}%
  \let\@fs@capt\floatc@ruled
  \def\@fs@pre{}%
  \def\@fs@mid{}%
  \def\@fs@post{}
  \let\@fs@iftopcapt\iffalse 
}
\makeatother

\newcommand{\orange}[1]{\textcolor{orange}{#1}}

\newcommand{\violet}[1]{\textcolor{violet}{#1}}
\newcommand{\darkGreen}[1]{\textcolor{darkGreen}{#1}}
\newcommand{\myblue}[1]{\textcolor{myblue}{#1}}
\newcommand{\darkblue}[1]{\textcolor{darkblue}{#1}}

\newcommand\blfootnote[1]{%
  \begingroup
  \renewcommand\thefootnote{}\footnote{#1}%
  \addtocounter{footnote}{-1}%
  \endgroup
}

\title{InternVideo-Next: Towards World Understanding Video Models}

\author{
    Chenting Wang$^{1,2}$\quad
    Yuhan Zhu$^{2,5}$\quad 
    Yicheng Xu$^{2}$\quad 
    Jiange Yang$^{2,5}$\quad \\
    Lang Lin$^{2}$ \quad 
    Ziang Yan$^{2}$\quad
    Yali Wang$^{2,4}$\quad
    Yi Wang$^{2,3}$\quad
    Limin Wang$^{1,2,5,\spadesuit}$\vspace{0.2em}\\
    \footnotesize{$^1$Shanghai Jiao Tong University}
    \quad $^2$Shanghai AI Laboratory
    \quad $^3$Shanghai Innovation Institute \\
    \footnotesize{$^4$Shenzhen Institutes of Advanced Technology, China \quad
    $^5$Nanjing University}
}
\begin{document}
\maketitle
\begin{abstract}
Large-scale video-text pretraining achieves strong performance but depends on noisy, synthetic captions with limited semantic coverage, often overlooking implicit world knowledge such as object motion, 3D geometry, and physical cues. In contrast, masked video modeling (MVM) directly exploits spatiotemporal structures but trails text-supervised methods on general tasks.
We find this gap arises from overlooked architectural issues: pixel-level reconstruction struggles with convergence and its low-level requirement often conflicts with semantics, while latent prediction often encourages shortcut learning.
To address these, we disentangle the traditional encoder-decoder design into an Encoder-Predictor-Decoder (EPD) framework, where the predictor acts as a latent world model, and propose InternVideo-Next, a two-stage pretraining scheme that builds a semantically consistent yet detail-preserving latent space for this world model.
First, conventional linear decoder in pixel MVM enforces the predictor’s output latent to be linearly projected to, thus separable in pixel space, causing the conflict with semantic abstraction.
Our Stage 1 proposes a conditional diffusion decoder and injects reliable image-level semantic priors to enhance semantics and convergence, thus bridging pixel-level fidelity with high-level semantic abstraction.
Stage 2 further learns world knowledge by predicting frozen Stage 1 targets within this space, mitigating shortcut learning.
Trained on public, unlabeled videos, InternVideo-Next achieves state-of-the-art results across benchmarks and provides a scalable path toward general video representation learning. Code and models will be released at \href{https://github.com/opengvlab/internvideo}{https://github.com/OpenGVLab/InternVideo}. 
\end{abstract}    
\section{Introduction}
\label{sec:intro}

\begin{figure}[t]
    \centering
    \includegraphics[width=\linewidth]{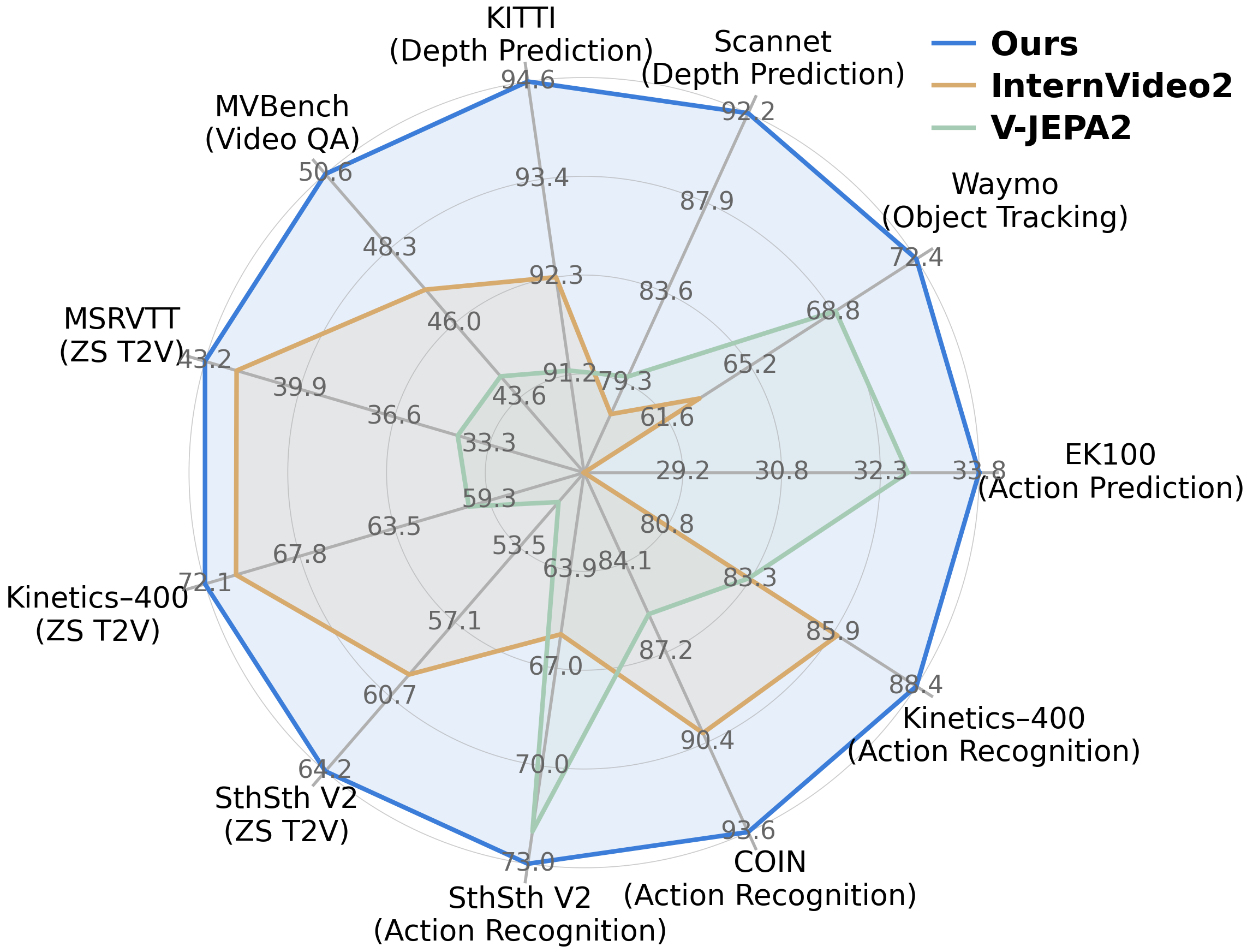}
    \vspace{-0.6cm}
    \caption{
        \textbf{Comparison with previous SOTA methods of size ViT-Large.} With only public sources, our model excels at general video benchmarks within a probing setting where ViTs are frozen to directly show representation's quality. The benchmarks involve scene-related, motion-related, complex video-language related and implicit world knowledge (3D geometric prior, causal relations and fine-grained object motion) related tasks.
    }
    \vspace{-0.5cm}
    \label{fig:figure1}
\end{figure}

\begin{figure*}[t]
    \centering
    \includegraphics[width=\textwidth]{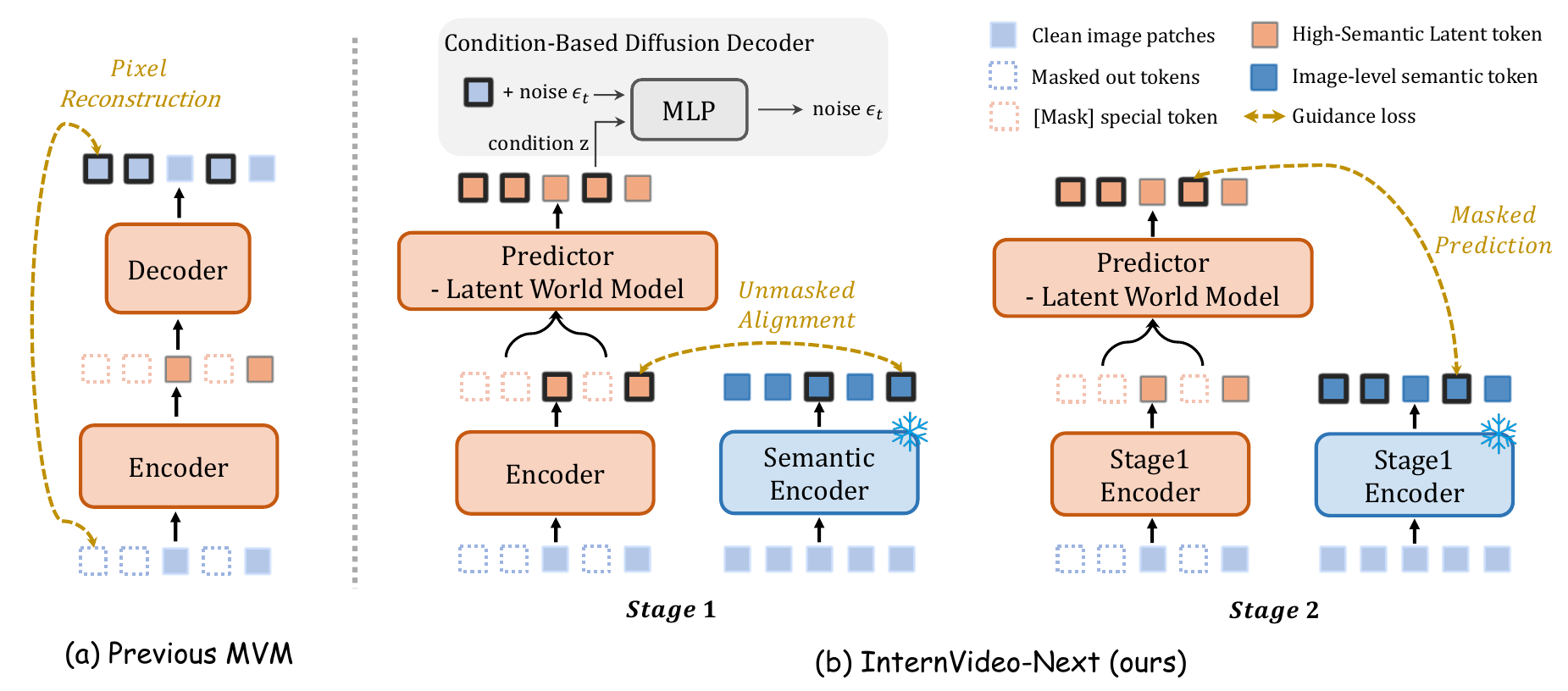}
    \vspace{-0.6cm}
    \caption{
        \textbf{Our two-stage self-supervised video pretraining framework.} For general video understanding and building real world-understanding video foundation models, we propose InternVideo-Next, which is simple, scalable, efficient and reproducible. 
    }
    \vspace{-0.5cm}
    \label{fig:videomaeplus}
\end{figure*}

\blfootnote{$\spadesuit$ Corresponding author (lmwang@nju.edu.cn).}

Video, as one of the most dynamic and information-rich modalities, offers a window into the physical world. It embeds not only spatiotemporal dynamics~\cite{tsn,tsm,slowfast,r3d} but also causal relationships, 3D geometrical priors and physical cues~\cite{scaling4drepresentations,vjepa2}. These are all essential components for building genuine video understanding models.
Such models are crucial for advancing embodied AI~\cite{vln,scalevln}, procedural reasoning~\cite{dinowm}, and the next generation of multimodal large language models~\cite{blip2, videochat, videochatflash, llava, liu2023improvedllava}. However, learning such knowledge from raw data, especially in a scalable and unbiased manner, remains a challenge.

Recent progress in large-scale video representation learning often falls into two categories. 
1) Text-supervised methods, leveraging large-scale video-text alignment~\cite{xu2021videoclip, umt, iv2, vprism, lu2019vilbert}, have achieved impressive performance on semantic, human-centric tasks like action recognition. However, due to expensive annotation costs, currently their reliance on noisy, synthetic annotations with limited semantic coverage introduces external biases, making it difficult to capture non-semantic vision features. They often fall short in tasks that require detailed object motion, causal structure or 3D geometrical priors. 
2) Self-supervised video pretraining~\cite{videomae, vjepa, vjepa2}, in contrast, holds the promise of learning directly from the spatiotemporal structure of videos, offering a path towards truly unbiased and scalable world modeling. However, existing methods still lag behind text-supervised counterparts on benchmarks requiring strong semantics about main subject like Kinetics-400~\cite{k400}. 

We contend that this performance gap of self-supervised video pretraining is not an intrinsic limitation, but rather stems from fundamental architectural challenges unsolved in current methods:
1) \textbf{Bridge pixel-level fidelity with high-level semantic abstraction} 
2) \textbf{Learn robust spatiotemporal dynamics, causal relations and 3D geometric priors from prediction without ``shortcut solutions"}. 

Popular MVM methods, like MAE~\cite{mae}, take a conventional Encoder-Decoder paradigm, where ViT-Decoder~\cite{vit} directly takes Encoder output and generates reconstructed pixels.
We explicitly decouple it into Predictor and Decoder, and consider the whole MVM paradigm as a new unified Encoder–Predictor–Decoder (EPD) overview. 
Such disentanglement enables a clearer inspection of the Predictor’s output latent space that is often neglected, which we find is essential in dealing with the unsolved challenges.
It also reveals a key insight overlooked by previous works: the Encoder and Predictor should share a semantically rich yet detail-faithful output latent space, making the predictor a possible Latent World Model. 
Enforcing this compels the Predictor to complete missing content using genuine spatiotemporal relationships and implicit world knowledge rather than trivial correlations, further boosting semantic abstraction with implicit world knowledge like geometry and motion in the Encoder's representation. 

In light of this, we propose InternVideo-Next, a novel two-stage video pretraining framework, as Figure~\ref{fig:videomaeplus}:

In \textbf{Stage~1}, we construct such a semantically grounded, detail-preserving and structurally consistent latent space by: 
1) leveraging a pixel reconstruction framework as foundation to learn fine-grained pixel details and spatiotemporal priors. 
2) using a condition-based diffusion generative decoder in place of the popular Linear Decoder head in turning Predictor output latent to pixels. Such a diffusion decoder is helpful to maintain high semantics in the Predictor output space as common linear decoder may require the output to be linearly separable to pixels, which is harmful to balance semantic information with fine-grained details.
3) integrating semantic priors from a pretrained Image semantic model. Unlike video–text pretraining, which is limited by the sparsity and noisiness of video captions, image–text pretraining benefits from massive, high-quality corpora whose captions more comprehensively describe visual content. Such priors are cleaner and can accelerate model's convergence by focusing on more temporal-centric information, meanwhile boosting multi-modal friendliness.

Building upon this, \textbf{Stage 2} then focuses on learning spatiotemporal dynamics and causal relations within this already coherent latent space. We employ a latent-space prediction objective, where a student model learns to predict the representations generated by a frozen teacher, where the two models are both initialized with weights learned in Stage 1. Crucially, the semantically rich and detail-preserving latent space established in Stage 1 prevents the ``shortcut solutions" that plague traditional latent-prediction methods, pushing the model to learn genuine predictive world knowledge rather than superficial temporal statistics.

As shown in \textbf{Figure~\ref{fig:figure1}}, \textbf{InternVideo-Next} achieves state-of-the-art results across a wide range of video understanding tasks. 
Remarkably, it is the \textbf{first video model trained without explicit video–text supervision} that surpasses video–text pretrained counterparts on both Kinetics-400 (action recognition) and SSv2 (fine-grained motion recognition). 
The learned representations also exhibit strong generalization to tasks requiring physical and 3D spatial intelligence, such as depth estimation and object-centric tracking, which are implicit world knowledge inherent in videos.
In addition, with lightweight LiT-style~\cite{zhai2022lit} fine-tuning, \textbf{InternVideo-Next} delivers competitive zero-shot video-text retrieval performance, highlighting our model representation's multimodal friendliness. 
Also, with preliminary exploration in chat-centric tasks in video, it holds the potential to be the foundation Encoder of next-generation multimodal video chat models. 
Our framework provides a unified and scalable pathway toward general-purpose world-understanding video models. All the code and model will be released upon publication.

\begin{table*}[t]
\centering
\setlength{\tabcolsep}{5pt}
\renewcommand{\arraystretch}{1.05}

\begin{minipage}[t]{0.33\linewidth}
\centering
\textbf{(a) Stage-1 module effect}\\[2pt]
\begin{tabular}{lcc}
\toprule
\textbf{Variant} & \textbf{K400} & \textbf{SSv2} \\
\midrule
Pixel Recon. Baseline & 47.2 & 28.1 \\
SigLIP Align only & 70.7 & 32.1 \\ \midrule
Pixel Rec. + Align & 69.8 & 31.8 \\
+ Diffusion Decoder & 74.2 & 35.4 \\
+ Text-Decoder Init & 69.4 & 31.3 \\
\rowcolor{blue!10}
\textbf{+ Keep Both} & \textbf{75.8} & \textbf{36.9} \\
\bottomrule
\end{tabular}
\end{minipage}
\hfill
\begin{minipage}[t]{0.33\linewidth}
\centering
\textbf{(b) Predictor size \& init}\\[2pt]
\begin{tabular}{lcc}
\toprule
\textbf{Predictor} & \textbf{K400} & \textbf{SSv2} \\
\midrule
ModernBert-L last2 & 73.2 & 34.8 \\
\rowcolor{blue!10}
\textbf{ModernBert-L last5} & \textbf{75.8} & \textbf{36.9} \\
last5 w/o init & 74.2 & 35.4 \\
ModernBert-L last8 & 75.6 & 37.0 \\ \midrule
Depth-5 ViT & 74.1 & 34.9 \\
Depth-12 ViT & 73.2 & 34.4 \\
\bottomrule
\end{tabular}
\end{minipage}
\hfill
\begin{minipage}[t]{0.33\linewidth}
\centering
\textbf{(c) Different Decoder settings}\\[2pt]
\begin{tabular}{lcc}
\toprule
\textbf{Decoder} & \textbf{K400} & \textbf{SSv2} \\
\midrule
Linear Head & 69.4 & 31.3 \\ 
3-Layer MLP Head & 69.7 & 31.2 \\ \midrule
DiffMLP D3 W1024 & 73.4 & 33.2 \\
\rowcolor{blue!10}
\textbf{DiffMLP D6 W1536} & \textbf{75.8} & \textbf{36.9} \\
DiffMLP D9 W2048 & 75.5 & 36.4 \\
\bottomrule
\end{tabular}
\end{minipage}

\vspace{0pt}
\caption{
\textbf{Ablations of Stage 1 design.} Linear probing results are shown.
(a) Adding semantic alignment boosts overall performance, while directly merging it into pixel reconstruction framework causes performance drop. Combining our diffusion decoder and text decoder augmentation make them fuse perfectly and results in accuracy gain. 
(b) Moderate predictor depth with initialization performs best in our scheme, outperforming popular results with Depth-12 ViT.
(c) D for number of MLP layers. and W for network width.
}
\vspace{-0.5cm}
\label{tab:ablation_stage1_maestyle}
\end{table*}

\section{Related Works}
\label{sec:related}

\paragraph{Text Supervised Pre-training.}

Clip-style pretraining has dominated current mainstream image pretraining frameworks~\cite{clip, siglip, siglip2}, benefiting from large-scale web-crawled image-text pair data. 
Video–text pretraining methods ~\cite{li2021learningspatiotemporalfeaturesvideo, xu2021videoclip, xclip, Cheng2022VindLUAR, vast} also achieve strong performance on benchmarks relying heavily on the main subject semantics like Kinetics-400.   
However, video captions are more difficult to be directly crawled from internet and are often generated by summarizing title and ASR information. Such synthetic annotations are typically noisier than images' and lack sufficient motion or spatial diversity, considering that videos are harder to fully describe than images.
As can be seen with Figure\ref{fig:figure1}, although video-text models excel at semantic-heavy tasks, they often fall short in tasks related to non-semantic information like depth, fine-grained motion and causal relation.
Our framework focuses more on unbiased learning from video itself, retaining high semantics, low-level details and world knowledge priors.

\paragraph{Masked Video Modeling.}

Masked video modeling (MVM) is the mainstream self-supervised video pretraining method, inspired by masked image modeling~\cite{mae, beit,beitv2,beit3}.  
VideoMAE~\cite{videomae} and MaskFeat~\cite{maskfeat} reconstruct masked video patches in the pixel domain, achieving strong performance but primarily capturing low-level appearance cues.  
Subsequent works such as VideoMAE-V2~\cite{videomaev2} improve masking strategies and decoder designs but still struggle to preserve high-level semantics.  
Latent-space prediction approaches, such as V-JEPA~\cite{vjepa}, move toward semantic abstraction by predicting feature representations instead of pixels, yet symmetric teacher–student structures often lead to shortcut learning or semantic drift. As seen in Figure~\ref{fig:figure1}, it falls short in appearance-intensive tasks, depth estimation tasks that require low-level details, and high-semantic tasks, meaning that such shortcut learning leads to missing details and prevents higher-level information.
Our approach unifies these paradigms under an Encoder–Predictor–Decoder (EPD) formulation and emphasizes maintaining semantic coherence between encoder and predictor latent spaces, combining the strengths of both.

\paragraph{The InternVideo Series.}

The InternVideo series~\cite{Wang2022InternVideoGV, iv2, fluxvit} has explored multiple ways in integrating unsupervised video learning with text alignment to bridge pixel-level fidelity with high-level semantic abstraction.  
InternVideo merges weights from pretrained VideoMAE and CLIP models using model ensemble, while InternVideo2 introduces a video-only stage that aligns unmasked representations from VideoMAE and CLIP encoders.  
These methods can be viewed as weight- or embedding-level fusion of video priors and language knowledge.  
However, they still can not fully solve the conflict between fine-grained details and high-level semantics.
InternVideo-Next revisits video-only pretraining from a task-level perspective by integrating CLIP-level semantic priors into an augmented video reconstruction framework, resolving such conflict.

\section{Method}
\label{sec:method}

\subsection{The EPD Disentanglement}

We formulate our unified Encoder–Predictor–Decoder (EPD) overview as follows: 
\textbf{$E$}: A Vision Transformer that extracts spatiotemporal representations from an input video.
\textbf{$P$}: A lightweight transformer that predicts latent representations for masked regions based on visible tokens. 
\textbf{$D$}: A reconstruction module that maps the predictor’s output latent to target space, either pixels or target latent.

\subsection{S1: Semantic-Guided Pixel Reconstruction}

\paragraph{Semantic Alignment Loss.}  
Our approach leverages semantic knowledge from the image domain via a frozen SigLIP encoder. This allows us to inject a strong image-level semantic prior into our video-only pre-training, offloading the need for noisy video-text annotations while focusing the model's learning on spatiotemporal dynamics:
\begin{equation}
\mathcal{L}_{\text{sem}} = -\cos\!\big(E(X_{\text{vis}}), \text{vis}(\text{SigLIP}(X))\big),
\end{equation}
We use a cosine similarity loss between the student’s embedding of a masked video and the corresponding visible region of the teacher’s embedding of the full video. 
Stage~1 jointly optimizes pixel reconstruction and semantic alignment with the same loss weight.

\paragraph{Semantic-Aware Masking.}  
Semantic mask prioritizes temporally informative regions using attention scores derived from the semantic teacher with a top-k selection.

\paragraph{Diffusion Decoder.}  
Unlike the popular linear decoder~\cite{mae}, we adopt a lightweight conditional diffusion~\cite{diffusion} decoder to model the distribution of each patch independently. The diffusion process and loss function follow ~\cite{diffimproved}, where noise schedule has a cosine shape, with 1000 steps at training time. We use a small MLP consisting of a few residual blocks~\cite{resnet} for denoising. The denoising MLP is conditioned on a vector z produced by the Predictor and outputs corresponding pixels. Since it only models a patch's latent distribution, a small MLP works perfectly and does not involve much overhead.

\paragraph{Text-Decoder Initialization.}  Conventional pixel reconstruction framework uses zero-init ViT. Our Predictor $P$ is initialized with weights from a pretrained text decoder~\cite{modernbert,lu2019vilbert}, which provides better semantic priors and the smooth translation of high semantics between the two latent spaces, and proved to demand fewer layers than common methods.

\subsection{S2: Semantically Coherent Latent Prediction}

Stage~2 builds upon the semantically aligned latent spaces obtained in Stage~1 to further enhance temporal abstraction and representation generalization.

\paragraph{Initialization.}  
Both the student and teacher networks are initialized with the weights learned in Stage 1. And the Stage 1 predictor is also kept in Stage 2.

\paragraph{Multi-Block Masking.}  
To strengthen temporal reasoning, we apply a multi-block masking~\cite{vjepa}, where large contiguous spatiotemporal blocks in a video are masked. This strategy increases prediction difficulty and mitigates information leakage, which is better for Stage 2's target of implicit world knowledge by prediction.

\paragraph{Latent Prediction Objective.}  
The student predicts the teacher’s latent representations for masked regions. This objective enforces consistency in the latent space without direct pixel reconstruction, enabling the model to focus on abstract semantic and temporal patterns.

\paragraph{Frozen Teacher Target.}  
Unlike V-JEPA, the teacher network is frozen with Stage 1 initialization to prevent shortcut learning, as the Stage 1 initialization is already detail-preserving and of high semantics. 

\begin{table}[t]
\centering
\setlength{\tabcolsep}{6pt}
\renewcommand{\arraystretch}{1.05}
\begin{tabular}{lccc}
\toprule
\textbf{Stage1-Variant} & \textbf{K400} & \textbf{SSv2} & \textbf{Scannet} \\
~ & Acc1$\uparrow$ & Acc1$\uparrow$ & $\delta_1\uparrow$ \\
\midrule
DinoV2 Align only & 68.4 & 29.5 & 43.3 \\
Clip-ViT Align only & 69.1 & 31.2 & 40.6 \\
SigLIP Align only & 70.7 & 32.1 & 42.1 \\
\midrule
\multicolumn{4}{l}{\darkGreen{\textit{InternVideo2 Stage1 Strategy}}} \\
SigLIP+VideoMAE align & 70.4 & 32.5 & 50.1 \\ \midrule
Ours \textit{w/ DinoV2 align} & 69.3 & 31.5 & 58.3 \\
Ours \textit{w/ Clip-ViT align} & 73.2 & 34.6 & 57.8 \\
\rowcolor{blue!10}
\textbf{Ours \textit{w/ SigLIP align}} & \textbf{75.8} & \textbf{36.9} & \textbf{59.4} \\
\bottomrule
\end{tabular}
\vspace{-1pt}
\caption{
\textbf{Comparison with different supervision signals.}
Also, we compare results with InternVideo2-S1, which directly aligns with two teachers as a first attempt to balance pixel-level detail and semantic abstraction, in fair training recipe. 
}
\label{tab:ablation_stage1_task_unification}

\vspace{10pt}

\setlength{\tabcolsep}{6pt}
\renewcommand{\arraystretch}{1.05}
\begin{tabular}{lcc}
\toprule
\textbf{Stage-2 Variant} & \textbf{K400} & \textbf{SSv2} \\
\midrule
Stage~1 & 75.8 & 36.9 \\ \midrule
\rowcolor{blue!10}
\textbf{Our Stage~2 setting} & \textbf{76.9} & \textbf{56.9} \\
\textit{replace w/} Zero-Init V-JEPA Predictor & 74.8 & 53.8 \\
\textit{replace w/} Momentum 0.9998 Target & 74.1 & 54.3 \\
\textit{replace w/} Frozen SigLIP2 Target & 75.4 & 45.7 \\
\textit{replace w/} Frozen InternVideo2 Target & 74.3 & 47.4 \\
\textit{if w/} Unmasked Token Alignment Loss & 75.7 & 51.1 \\
\textit{if w/} Pixel Reconstruction Loss & 76.8 & 57.0 \\
\bottomrule
\end{tabular}
\vspace{-1pt}
\caption{
\textbf{Stage-2 ablation.}
Our full configuration yields the strongest temporal abstraction and overall accuracy.
}
\vspace{-0.5cm}
\label{tab:ablation_stage2_variants}
\end{table}

\section{Experiments}

\paragraph{Implementation.}
For our pretraining, we utilize the last five layers of ModernBert-Large~\cite{modernbert} as our Predictor. We use SigLIP2-1B~\cite{siglip2} as our semantic teacher in our final version, and use SigLIP2-Large in the Ablation study. 
For Stage~1 pre-training, we use a mask ratio of 80\% and a learning rate of 1e-3. For ablation study, we use 32 A100 with a batch size of 1024 for 30 epochs in both Stage 1 and Stage 2. And for final training, we use 64 A100 with a batch size of 2048 for 50 epochs in Stage 1 and 100 epochs in Stage 2. For more, please refer to the Supplement.

\paragraph{Datasets.}
Unless otherwise stated, we use K-Mash~\cite{iv2} for training, including 1.1M video data from K400~\cite{k400}, K600~\cite{k600}, K700~\cite{k700}, SSv2~\cite{goyal2017something}, ANet~\cite{activitynet}, HACS~\cite{hacs}, and MiT~\cite{mit} with duplicates removed. For ablation, we use a subset K710 containing only the Kinetics datasets.

\subsection{Ablation Study}
We conduct comprehensive ablations to analyze performance of our training framework on appearance-based, motion-centric and depth-estimation benchmarks, including K400~\cite{k400}, SSv2~\cite{goyal2017something}, and ScanNet~\cite{scannet}. We leverage Linear Probing results on action recognition tasks for clear presentation of representation quality improvement. Results on ScanNet are based on probing results using learnable queries and a single-layer trainable cross-attention head.

\paragraph{Effects of Stage~1 Settings.}
Table~\ref{tab:ablation_stage1_maestyle} presents detailed ablations on the Stage~1 design. 
As shown in (a), the pixel reconstruction baseline achieves 47.2\% on K400, confirming its limited semantic abstraction ability.  
Using SigLIP alignment alone without reconstruction boosts recognition accuracy.  
However, naively combining pixel reconstruction and alignment hurts overall performance due to optimization conflicts. Introducing the \emph{diffusion decoder} reverses such degradation into a +4.4\% performance gain, demonstrating strong complementarity between the two supervision signals, if not hurt by the original Linear Decoder. It highlights the effect of our EPD disentanglement, which provides a novel reconsideration and inspires us to re-use the now seldom-utilized pixel reconstruction framework.
(b) studies the impact of predictor architecture and initialization. The vanilla ViT predictor without initialization, as used in the original setting, underperforms our solution.

\begin{table}[t]
\centering
\setlength{\tabcolsep}{5pt}
\renewcommand{\arraystretch}{1.05}
\begin{tabular}{lccccc}
\toprule
\textbf{Stage} & \textbf{Mask Type} & \textbf{\#F} &
\multicolumn{1}{c}{\textbf{K400}} &
\multicolumn{1}{c}{\textbf{SSv2}} &
\multicolumn{1}{c}{\textbf{Scannet}} \\
 &  &  & Acc1$\uparrow$ & Acc1$\uparrow$ & $\delta_1\uparrow$ \\
\midrule
\multicolumn{6}{l}{\darkGreen{\textit{Effect of Masking Strategy (ViT-B)}}} \\
\midrule
~ & \cellcolor{blue!10}\textbf{Semantic} & \cellcolor{blue!10}\textbf{8} & \cellcolor{blue!10}\textbf{75.8} & \cellcolor{blue!10}\textbf{36.9} & \cellcolor{blue!10}\textbf{59.4} \\
\multirow{-2}{*}{Stage-1} & Multi-block & 8 & 74.4 & 36.3 & 58.1 \\ \midrule
~ & Semantic & 8 & 75.2 & 52.3 & 61.1 \\
\multirow{-2}{*}{Stage-2} & \cellcolor{blue!10}\textbf{Multi-block} & \cellcolor{blue!10}\textbf{8} & \cellcolor{blue!10}\textbf{76.9} & \cellcolor{blue!10}\textbf{56.9} & \cellcolor{blue!10}\textbf{66.1}\\
\midrule
\multicolumn{6}{l}{\darkGreen{\textit{Effect of Temporal Frame Count (ViT-B)}}} \\
\midrule
 & ~ & 8  & 75.8 & 36.9 & 59.4\\
 &  & \cellcolor{blue!10}\textbf{16} & \cellcolor{blue!10}\textbf{76.4} & \cellcolor{blue!10}\textbf{38.4} & \cellcolor{blue!10}\textbf{59.8}\\
\multirow{-3}{*}{Stage-1} & \multirow{-3}{*}{Semantic} & 32 & 76.6 & 38.6 &  59.3\\ \midrule
~ & ~ & 8  & 77.0 & 57.5 & 67.1 \\
 &  & 16 & 77.4 & 58.4 & 68.7 \\
\multirow{-3}{*}{Stage-2} & \multirow{-3}{*}{Multi-block} & \cellcolor{blue!10}\textbf{32} & \cellcolor{blue!10}\textbf{78.1} & \cellcolor{blue!10}\textbf{59.4} & \cellcolor{blue!10} \textbf{70.1}\\
\bottomrule
\end{tabular}
\vspace{-1pt}
\caption{
\textbf{Impact of masking and frame settings.} Each Stage-2 model is based on the highlighted Stage-1 model in the same table. 
}
\vspace{-0.3cm}
\label{tab:ablation_mask_frame}
\end{table}

\begin{table*}[tp]
    \centering
    \begin{tabular}{l|l|c|c|c|c|c}
        \Xhline{1.0pt}
        \textbf{Model Name} & \textbf{ViT Size} & \textbf{Data} & \textbf{GPU-hrs} & \textbf{K400} $\uparrow$ & \textbf{SSv2} $\uparrow$ & \textbf{Coin} $\uparrow$  \\
        \Xhline{1.0pt}
        \multicolumn{6}{@{}l}{\darkGreen{\textit{Image Models}}} \\
        SigLIP2~\cite{siglip2} & Giant & 10B & - & 85.6 & 47.9 & 91.8 \\
        DinoV2~\cite{dino} & Giant & 142M & - & 83.1 & 50.0 & 87.8 \\
        \Xhline{1.0pt}
        \multicolumn{6}{@{}l}{\darkGreen{\textit{Models Trained with Video-Text Pairs}}} \\
        InternVideo2$_{s2}$~\cite{iv2} & Base & 25.5M & - & 84.9 & 64.7 & 88.7 \\
        InternVideo2$_{s2}$~\cite{iv2} & Large & 25.5M & - & 86.0 & 65.9 & 90.1 \\
        InternVideo2$_{s2}$~\cite{iv2} & 1B & 25.5M & 30K & 87.9 & 67.3 & 91.7  \\
        InternVideo2$_{s2}$~\cite{iv2} & 6B & 400M & 200K & 88.8 & 67.7 & 92.6 \\
        VideoPrism~\cite{vprism} & Base & 618M & - & 84.2 & 63.6 & - \\
        VideoPrism~\cite{vprism} & 1B & 618M & 250K$^*$ & 87.2 & 68.5 & - \\
        \Xhline{1.0pt}
        \multicolumn{6}{@{}l}{\darkGreen{\textit{Models Trained with Video Data Only}}} \\
        VideoMAEv2~\cite{videomaev2} & Large & 1.35M & - & 80.9 & 54.9 & 83.2 \\
        VideoMAEv2~\cite{videomaev2} & 1B & 1.35M & 15K & 82.8 & 56.1 & 84.6 \\
        V-JEPAv1~\cite{vjepa} & Large & 2M & 8K$^*$ & 80.8 & 69.5 & 83.0 \\
        V-JEPAv2~\cite{vjepa2} & Large & 22M & 10K$^*$ & 83.3 & 72.0 & 85.9 \\
        InternVideo2$_{s1}$~\cite{iv2} & 1B & 1.1M & 15K & 85.6 & 57.4 & 89.6 \\
        InternVideo2$_{s1}$~\cite{iv2} & 6B & 2.1M & 110K & 86.0 & 59.0 & 90.3 \\
        \rowcolor{blue!10}
        \textbf{InternVideo-Next$_{s1}$} & \textbf{Base} & \textbf{1.1M} & 1.5K &  \textbf{84.8} & \textbf{58.6} & \textbf{89.9} \\
        \rowcolor{blue!10}
        \textbf{InternVideo-Next$_{s2}$} & \textbf{Base} & \textbf{1.1M} & 3.4K &  \textbf{85.9} & \textbf{70.1}& \textbf{91.4} \\
        \rowcolor{blue!10}
        \textbf{InternVideo-Next$_{s1}$} & \textbf{Large} & \textbf{1.1M} & 3.6K &  \textbf{87.1} & \textbf{65.5} & \textbf{92.0}  \\
        \rowcolor{blue!10}
        \textbf{InternVideo-Next$_{s2}$} & \textbf{Large} & \textbf{1.1M} & 9.7K &  \textbf{88.4} & \textbf{73.0} & \textbf{93.6} \\
        \Xhline{1.0pt}
    \end{tabular}
    \caption{\textbf{Attentive Probing Results on Video Action Recognition Datasets K400, SSv2 and COIN.} We report top@1 results with a single probing head trainable. The training cost is estimated with equivalent A100 GPU Hours. $^*$ means estimated result, as not included in the paper. Models are tested with 16 frames, and with 1 clips $\times$3 crops. Our models achieve the best performance across model sizes and with the smallest, public and annotation-free training data, highlighting our method's efficiency.}
    \label{tab:action_recognition}
    \vspace{-0.5cm}
\end{table*}

Panel (c) compares different decoder designs. A lightweight linear head underfits complex spatial relations, while our diffusion-based MLP decoders progressively improve as capacity grows. The medium-sized \texttt{Depth 6, Width 1536} variant achieves the best trade-off between representation richness and computational cost. 

\paragraph{Different supervision for Stage 1's semantic prior.}
Table~\ref{tab:ablation_stage1_task_unification} reveals that our framework shows generalizable performance gain with different semantic alignment targets, whether on common recognition tasks or 3D understanding tasks with low-level details. Supervision from semantic image models performs best. It also compares our unified Stage~1 strategy with the multi-teacher alignment scheme in InternVideo2, which aligns with both CLIP and VideoMAE to balance pixel-level detail and semantic abstraction.

\paragraph{Effects of Stage~2 Settings.}
Stage~2 further boosts temporal abstraction and causal relations via masked latent prediction.  
As shown in Table~\ref{tab:ablation_stage2_variants}, our configuration achieves the strongest results.  
Replacing our predictor with a zero-initialized V-JEPA variant, unfreezing the target encoder or using other initialization of the target encoder all lead to degradation, confirming the importance of semantically coherent initialization and a stable target that encodes both high semantics and pixel-level details.  
Adding unmasked token alignment as in Stage~1 introduces noise to latent targets and harms motion modeling.  
Pixel-space reconstruction during this stage provides marginal benefit to SSv2, as the targets, produced by our Stage~1 encoder, already contain enough pixel-level details.

\paragraph{Masking Strategy and Temporal Window.}
Table~\ref{tab:ablation_mask_frame} analyzes masking and temporal length configurations.  
Semantic masking in Stage~1 helps the model attend to spatially meaningful regions and bridge high-level semantic abstraction with pixel-level fidelity. 
Multi-block masking in Stage~2 makes the prediction task harder and forces robust spatiotemporal dynamics and causal relations learned in the Predictor. 
Moreover, increasing the number of input frames consistently enhances performance, demonstrating that richer temporal diversity strengthens temporal abstraction. We finally chose F16 for Stage~1 and F32 for Stage~2 for balance between accuracy and efficiency.

\begin{figure*}[t]
    \centering
    \includegraphics[width=\textwidth]{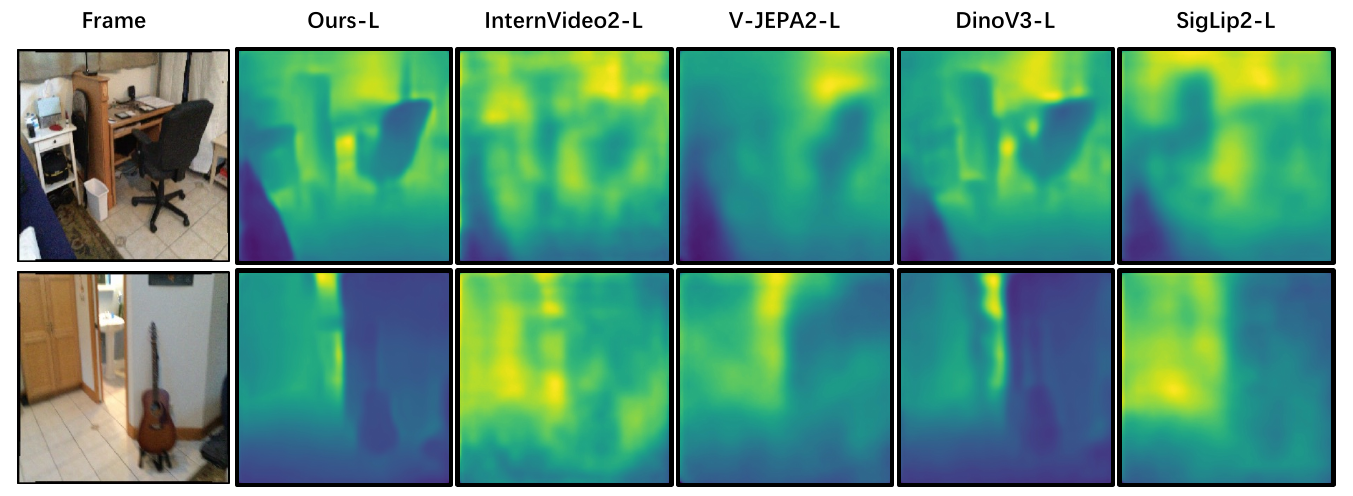}
    \vspace{-0.7cm}
    \caption{
        \textbf{Video Depth estimation visualization.} All models are trained with a frozen setting with only a probing head (VDA head from Video Depth Anything~\cite{videodepthanything}) module trainable. Graphs are from the first video of the ScanNet Dataset. Our model shows potential for building next-generation 3D-spatial intelligence models from video sources. Videos are resized with a spatial resolution of 224.
    }
    \vspace{-0.3cm}
    \label{fig:depth_visual}
\end{figure*}

\begin{table}[t]
    \centering
    \setlength\tabcolsep{2pt}
    \resizebox{0.94\linewidth}{!}{
        \begin{tabular}{l|cc|cc}
            \Xhline{1.0pt}
            \multirow{2}{*}{\textbf{Model Name}} & \multicolumn{2}{c|}{\textbf{Scannet}} & \multicolumn{2}{c}{\textbf{KITTI}} \\
            \cmidrule(r){2-3} \cmidrule(l){4-5}
            ~ & ARel$\downarrow$ & $\delta_1$$\uparrow$ & ARel$\downarrow$ & $\delta_1$$\uparrow$ \\
            \Xhline{1.0pt}
            \multicolumn{5}{@{}l}{\darkGreen{\textit{SOTA solution with VDA head and specific training}}} \\
            VideoDepthAnything~\cite{videodepthanything} & 8.7 & 92.6 & 8.3 & 94.6 \\
            \Xhline{1.0pt}
            \multicolumn{5}{@{}l}{\darkblue{\textit{Image Models \textit{w/} a simple probing head~\cite{scaling4drepresentations}}}} \\
            SigLip2-L~\cite{siglip2} & 26.4 & 50.4 & 16.8 & 74.9 \\
            DinoV3-L~\cite{dinov3} & 24.0 & 59.1 & 13.5 & 81.8 \\
            \Xhline{1.0pt}
            \multicolumn{5}{@{}l}{\darkblue{\textit{Video Models \textit{w/} a simple probing head~\cite{scaling4drepresentations}}}} \\
            VideoMAEv2-L\cite{videomaev2} & 19.3 & 66.9 & 11.9 & 84.6 \\
            InternVideo2-L~\cite{iv2} & 23.4 & 57.0 & 12.9 & 83.0 \\
            V-JEPA2-L~\cite{vjepa2} & 20.9 & 63.1 & 10.9 & 86.8 \\
            \rowcolor{blue!10}
            \textbf{InternVideo-Next-L} & \textbf{13.7} & \textbf{81.5} & \textbf{9.9} & \textbf{88.6} \\
            \Xhline{1.0pt}
            \multicolumn{5}{@{}l}{\darkGreen{\textit{Image Models \textit{w/} VDA head with temporal fusion~\cite{videodepthanything}}}} \\
            SigLip2-L~\cite{siglip2} & 19.9 & 65.6 & 12.2 & 84.2 \\
            DinoV3-L~\cite{dinov3} & 9.6 & 91.2 & 7.0 & 93.0 \\
            \Xhline{1.0pt}
            \multicolumn{5}{@{}l}{\darkGreen{\textit{Video Models \textit{w/} VDA head with temporal fusion~\cite{videodepthanything}}}} \\
            VideoMAEv2-B~\cite{videomaev2} & 21.8 & 60.6 & 9.6 & 89.7 \\
            InternVideo2-B~\cite{iv2} & 15.7 & 76.2 & 8.8 & 91.3 \\
            \rowcolor{blue!10}
            \textbf{InternVideo-Next-B} & \textbf{11.3} & \textbf{87.3} & \textbf{7.5} & \textbf{93.2} \\
            \Xhline{1.0pt}
            VideoMAEv2-L~\cite{videomaev2} & 17.9 & 70.4 & 9.5 & 89.7 \\
            InternVideo2-L~\cite{iv2} & 15.1 & 77.8 & 8.1 & 92.3 \\
            V-JEPA2-L~\cite{vjepa2} & 14.4 & 79.6 & 8.7 & 91.2 \\
            \rowcolor{blue!10}
            \textbf{InternVideo-Next-L} & \textbf{9.2} & \textbf{92.2} & \textbf{6.7} & \textbf{94.6} \\
            \Xhline{1.0pt}
        \end{tabular}
    }
    \caption{\textbf{Probing Results on Video Depth-Estimation Datasets Scannet and KITTI.} Our InternVideo-Next with such a simple probing setting can achieve comparable results with the carefully designed and trained Video Depth Anything model. ARel for `absolute relative error' and $\delta_1$ for `relative accuracy'~\cite{videodepthanything}. }
    \label{tab:depth_estimation}
    \vspace{-0.5cm}
\end{table}

\begin{table}[t]
    \centering
    \setlength\tabcolsep{2pt}
    \resizebox{0.65\linewidth}{!}{
        \begin{tabular}{l|c}
            \Xhline{1.0pt}
            \multirow{2}{*}{\textbf{Model Name}} & \textbf{Waymo Open} \\
            ~ & mean IOU$\uparrow$ \\
            \Xhline{1.0pt}
            \multicolumn{2}{@{}l}{\darkGreen{\textit{Image Models}}} \\
            SigLip2-L~\cite{siglip2} & 52.3 \\
            DinoV3-L~\cite{dinov3} & 59.7 \\
            \Xhline{1.0pt}
            \multicolumn{2}{@{}l}{\darkGreen{\textit{Video Models}}} \\
            InternVideo2-L~\cite{iv2} & 63.0 \\
            V-JEPA2-L~\cite{vjepa2} & 68.9 \\
            \rowcolor{blue!10}
            \textbf{InternVideo-Next-L} & \textbf{72.4} \\
            \Xhline{1.0pt}
        \end{tabular}
    }
    \caption{\textbf{Probing Results on Video Object Tracking Task Waymo Open.} We follow the task setting in Scaling4D~\cite{scaling4drepresentations} to test representation's capability in capturing object-level motion.
    }
    \label{tab:object_tracking}
    \vspace{-0.3cm}
\end{table}

\subsection{Single-modality Tasks}
We evaluate our method on several single-modality tasks: including the \textit{Kinetics-400}~\cite{k400}, \textit{Something-Something V2}~\cite{goyal2017something} for action recognition, \textit{COIN}~\cite{coin} for long video classification, \textit{ScanNet}~\cite{scannet} and \textit{KITTI}~\cite{kitti} for monocular depth estimation, \textit{WayMo}\cite{waymo} for object tracking and \textit{EK100}\cite{epic_kitchen} for action prediction.

\paragraph{Video Classification.}
Table \ref{tab:action_recognition} reports probing results with different frozen backbones on general video classification tasks. We test the model in an `Attentive Probing' setting where the encoders are frozen and a single-layer attention pooling head is trained. Such Frozen Encoder settings can test representation's quality in an unbiased way. Our methods achieve the best results with only public data and less computation cost on these foundation tasks.

\paragraph{Depth Estimation.}
As discussed above, 3D geometric prior is also a part of the world knowledge that can be acquired from videos. Table~\ref {tab:depth_estimation} tests the Encoder's probing performance on 3D-related Monocular Video Depth Estimation tasks Scannet~\cite{scannet} and KITTI\cite{kitti} in two settings. One uses a single-layer cross-attention based Probing Head with learnable queries, as to directly show model's spatiotemporal embedding quality regarding Video Depth. Our InternVideo-Next achieves far better results than image models as it learns more direct 3D spatial information in the backbone.
And another uses a complex VDA head designed specifically for Image models with additional temporal fusion modules~\cite{videodepthanything}. Our InternVideo-Next excels at both settings, even achieving nearly SoTA performance in such a `Probing Setting' compared with the specifically designed and trained Video Depth Anything~\cite{videodepthanything} model. Detailed test settings are left in the Supplement.

\paragraph{Object Tracking.}
We follow the setting in Scaling 4D representations~\cite{scaling4drepresentations} to test model embedding's quality in capturing object-level motion with Object Tracking task on Waymo Open~\cite{waymo}. The task requires the model to predict  Bounding Box given the coordinates of an object in the first frame. We use learnable queries and a single-layer cross-attention pooling head with the frozen ViT backbone for testing. We will leave more details in the Supp. Our model achieves the best result in capturing object-level motion.

\paragraph{Video Prediction.}
Table \ref{tab:ek100} shows video prediction results on Epic-Kitchens-100~\cite{epic_kitchen}. The model predicts the future action given a contextual video clip. The outputs of the predictor and encoder are concatenated along the token dimension and fed to an attentive probe layer and three classification heads. We follow most of the settings in V-JEPA2~\cite{vjepa2} and report the mean-class recall-at-5 result. Our model performs better at predicting future actions.

\subsection{Multi-modality Tasks}
In this section, we evaluate the potential of our model in multi-modal learning scenarios through two representative experimental setups. 
First, following the LiT~\cite{zhai2022lit} methodology, we freeze the ViT backbone and train only a text encoder, initialized with MobileCLIP-B~\cite{vasu2024mobileclip}, on 25.5M~\cite{bain2021frozen, cc3m, cc12m, coco, vg, sbu, smit} publicly available video-text pairs from UMT~\cite{umt} for 5 epochs. This setup evaluates the model's zero-shot video-text retrieval performance, serving as a direct probe into the completeness and correctness of the visual embeddings in the semantic space. We choose InternVideo2$_{s2}$ and V-JEPA2 as our baseline model.
Second, to assess the model's applicability as a video foundation model for high-level dialogue tasks, we connect the frozen ViT with a frozen large language model, Qwen2-7B~\cite{wang2024qwen2}, via a trainable MLP connector. Only the MLP is fine-tuned. This is consistent with the common Stage~1 training protocol for video-based chat models to validate the compatibility and transferability of our visual representations in a generative video-LLM framework.

\begin{table}[t]
    \centering
    \setlength\tabcolsep{2pt}
    \resizebox{0.75\linewidth}{!}{
        \begin{tabular}{l|c|c|c}
        \Xhline{1.0pt}
        \textbf{Method} & \textbf{Verb} & \textbf{Noun} & \textbf{Action} \\
        \Xhline{1.0pt}
        \multicolumn{4}{@{}l}{\darkGreen{\textit{MLLM MCQ result}}} \\
        VideoLLaMa-7B~\cite{videollama} & 52.9 & 52.0 & 26.0 \\
        \Xhline{1.0pt}
        \multicolumn{4}{@{}l}{\darkGreen{\textit{Frozen Backbone Evaluation}}} \\
        V-JEPA2-L~\cite{vjepa2} & 57.8 & 53.8 & 32.7 \\
        \rowcolor{blue!10}
        \textbf{InternVideo-Next-L} & \textbf{58.9} & \textbf{56.4} & \textbf{34.0} \\
        \Xhline{1.0pt}
    \end{tabular}
    }
    \caption{
    \textbf{Results of Video Action Prediction task EK100.} We report mean-class recall-at-5 for verb, noun and action on the validation set of EK100, following the test pipeline in V-JEPA2.
    }
    \label{tab:ek100}
    \vspace{-0.cm}
\end{table}

\newcommand{\performanceIncrease}[1]{
  \textbf{\darkGreen{$\uparrow$\footnotesize{#1}}}
}

\newcommand{\performanceRelativeIncrease}[1]{
  \textbf{\darkblue{$\uparrow$\footnotesize{#1}}}
}

\begin{table}[t]
    \centering
    \setlength\tabcolsep{2pt}
    \resizebox{0.92\linewidth}{!}{
        \begin{tabular}{l|c|c|c|c}
        \Xhline{1.0pt}
        \textbf{Method} & \textbf{K400} & \textbf{K700} & \textbf{SSv2-MC} & \textbf{MiTv1} \\
        \Xhline{1.0pt}
        InternVideo2$_{s2}$-B~\cite{iv2} & 67.7 & 57.9 & 55.9 & 27.9 \\
        \rowcolor{blue!10}
        \textbf{InternVideo-Next-B} & \textbf{68.9} & \textbf{59.0} & \textbf{61.2} & \textbf{29.0} \\
        \hline
        V-JEPA2-L~\cite{vjepa2} & 60.2 & 51.4 & 53.0 & 25.4 \\
        InternVideo2$_{s2}$-L~\cite{iv2} & 70.7 & 61.9 & 59.6 & 30.6 \\
        \rowcolor{blue!10}
        \textbf{InternVideo-Next-L} & \textbf{72.1} & \textbf{63.0} & \textbf{64.2} & \textbf{32.0} \\
        \Xhline{1.0pt}
    \end{tabular}
    }
    \caption{\textbf{Results of Zero-shot Action Recognition on K400, K700, SSv2-MC and MiT.} Both models use a MobileCLIP-B text encoder and train with ViT frozen to test their representations' separability, alignment and completeness in text space.
    }
    \label{tab:more_retrieval_ar}
    \vspace{-0.2cm}
\end{table}

\paragraph{Zero-shot Action Recognition.}
Table \ref{tab:more_retrieval_ar} shows results on zero-shot action recognition tasks including K400~\cite{k400}, K700~\cite{k700}, SSv2~\cite{goyal2017something} and MiTv1~\cite{mit}. Our model outperforms InternVideo2$_{s2}$, which is the previous SoTA encoder for multimodal retrieval tasks, especially on motion-intensive task SSv2-Multiple Choice.

\paragraph{Zero-shot Retrieval.}
Table \ref{tab:t2v_retrieval} shows results on zero-shot text-to-video retrieval tasks, including MSRVTT~\cite{msrvtt}, DiDeMo~\cite{didemo}, ActivityNet~\cite{anet}, LSMDC~\cite{lsmdc} and MSVD~\cite{msvd}. Our model shows comparable performance. This highlights our model's embedding of the video is highly aligned with the semantic text space implicitly.

\paragraph{Chat-centric Tasks.}
We further evaluate our model's potential as foundation Encoder for generative video-LLM frameworks. We follow the recipe of VideoChat-Flash~\cite{videochatflash} Stage~1 and utilize a large-scaled train-set including LLava-558K\cite{llava}, S-MiT~\cite{smit}, 700k filtered subset of WebVid-10M~\cite{bain2021frozen}, VidLN\cite{vidln}, and SSv2-open-ended~\cite{goyal2017something} to fully train the connector. Such `Linear Probing' setting is widely used in the Stage-1 training of multi-modal chat models. Our model achieves better results on general spatiotemporal perception Video QA tasks MVBench~\cite{mvp} and Perception Test~\cite{perception_test}, and caption task Dream1K~\cite{tarsier}.

\begin{table}[t]
    \centering
    \setlength\tabcolsep{2pt}
    \resizebox{\linewidth}{!}{
        \begin{tabular}{l|c|c|c|c|c}
        \Xhline{1.0pt}
        \textbf{Method} & \textbf{MSR} & \textbf{DDM} & \textbf{ANet} & \textbf{LSMDC} & \textbf{MSVD} \\
        \Xhline{1.0pt}
        \multicolumn{6}{@{}l}{\darkGreen{\textit{Popular methods with full video-text training}}} \\
        VINDLU-L ~\cite{Cheng2022VindLUAR} & 32.0 & 36.9 & 30.9 & - & - \\
        InternVideo-L ~\cite{Wang2022InternVideoGV} & 40.7 & 31.5 & 30.7 & 17.6 & 43.4 \\ 
        UMT-L ~\cite{umt} & 40.7 & 48.6 & 41.9 & 24.9 & 49.0 \\
        ViClip-L ~\cite{internvid} & 42.4 & 18.4 & 15.1 & 20.1 & 49.1 \\
        LanguageBind-L ~\cite{zhu2023languagebind} & 42.8 & 39.7 & 38.4 & - & 54.1 \\
        \Xhline{1.0pt}
        \multicolumn{6}{@{}l}{\darkGreen{\textit{Frozen ViT with LiT Training using MobileCLIP-B}}} \\
        V-JEPA2-L~\cite{vjepa2} & 34.4 & 36.3 & 35.7 & 19.2 & 40.1 \\
        InternVideo2$_{s2}$-L~\cite{iv2} & 42.1 & 42.8 & 43.6 & 21.4 & 44.5 \\
        \rowcolor{blue!10}
        \textbf{InternVideo-Next-L} & \textbf{43.2} & \textbf{43.7} & \textbf{43.4} &\textbf{20.8} & \textbf{46.1} \\
        \Xhline{1.0pt}
    \end{tabular}
    }
    \vspace{-0.1cm}
    \caption{
    \textbf{Results of zero-shot text-to-video retrieval on MSRVTT, DiDeMo, ActivityNet, LSMDC and MSVD.} We only report the T2V R@1 accuracy here.
    }
    \label{tab:t2v_retrieval}
    \vspace{-0.1cm}
\end{table}

\begin{table}[tp]
    \centering
    \setlength\tabcolsep{2pt}
    \resizebox{0.95\linewidth}{!}{
        \begin{tabular}{l|c|c|c}
            \Xhline{1.0pt}
            \textbf{Encoder} & \textbf{MVBench} & \textbf{Percept Test} & \textbf{Dream1k} \\ \Xhline{1.0pt}
            Clip-L~\cite{clip} & 45.6 & 45.1 & 28.4 \\  
            SigLIP$_{336}$-L~\cite{siglip} & 46.7 & 44.1 & 29.2 \\
            SigLIP2$_{336}$-L~\cite{siglip2} & 46.9 & 45.8 & 29.6 \\ \Xhline{1.0pt}
            UMT-L~\cite{umt} & 45.0 & 44.5 & 24.6 \\ 
            V-JEPA2-L~\cite{umt} & 44.3 & 44.2 & 24.3 \\ 
            InternVideo2-L~\cite{iv2} & 47.0 & 46.7 & 28.7 \\ 
            \rowcolor{blue!10}
            \textbf{InternVideo-Next-L} & \textbf{50.6} & \textbf{49.2} & \textbf{29.8} \\ 
            \Xhline{1.0pt}
        \end{tabular}
    }
    \caption{\textbf{Results on Chat-Centric benchmarks MVbench~\cite{mvp}, Perception Test~\cite{perception_test} for General Perception and Dream1k~\cite{tarsier} for Detailed Caption.} Models are trained in a multi-modal `linear prob' setting where both the Encoder and the LLM are frozen, which is a common setting in the Stage~1 training for multi-modal chat models. For Dream1k we report the F1-score.}
    \label{tab:flux_chat}
    \vspace{-0.2cm}
\end{table}
\section{Conclusion and Future Work}
In this paper, we present InternVideo-Next, a two-stage video pretraining framework without video-text supervision that unifies pixel-level reconstruction and latent prediction under a coherent semantic space and learns by prediction with such a space. By enforcing semantic alignment between encoder and predictor, our method learns structured spatiotemporal priors from raw videos, surpassing previous video-only and video–text methods on general video tasks and proving InternVideo-Next's capability as a strong general video foundation model. 
For future works, we plan to investigate the scalability of this method, and extend this framework toward multi-modal and interactive world modeling, exploring efficient scaling and open-world generalization for more versatile video foundation models. 
\appendix

\begin{figure*}[h]
    \centering
    \includegraphics[width=\textwidth]{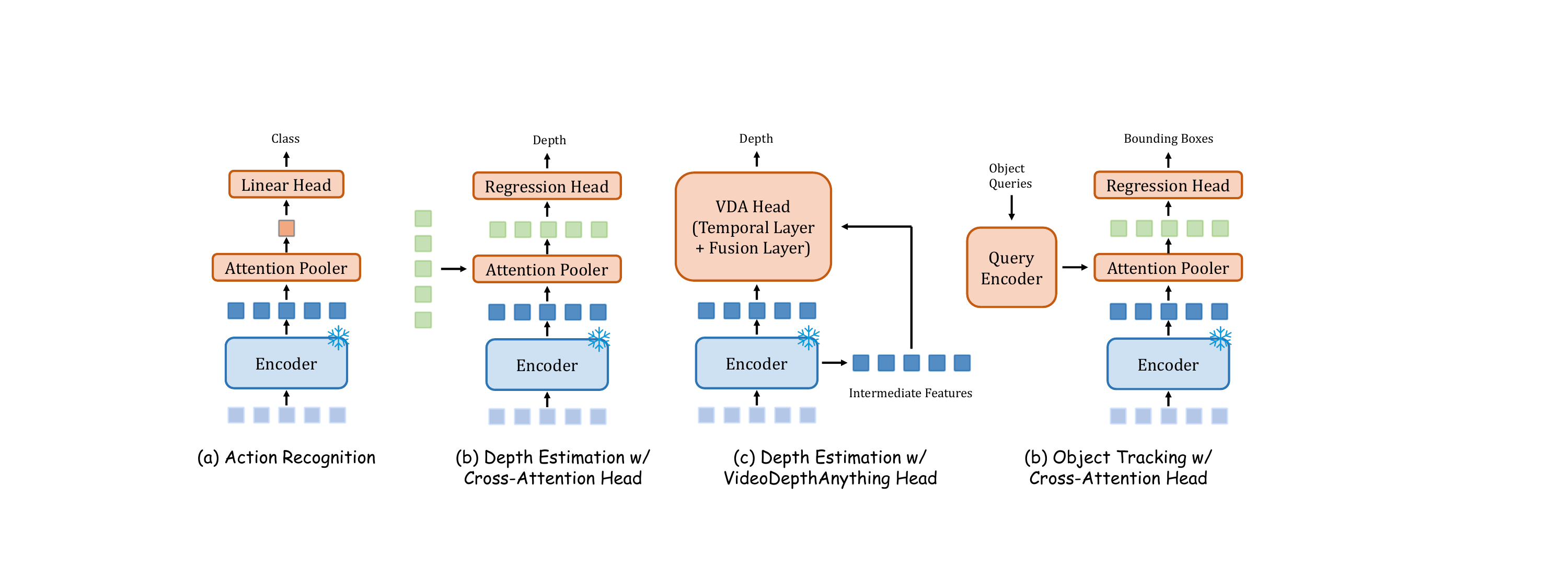}
    \caption{
        \textbf{Explanations of different probing settings in our paper.}
    }
    \label{fig:supp_main}
\end{figure*}

\section{More implementation details}

In this section, we introduce the model architectures, training hyperparameters, and test settings in our experiments.

\subsection{Model architecture and training details}

We use a patch of size 14 for our InternVideo-Next models. We report the detailed architecture of a Base-scaled model in Table \ref{tab:model_architecture}, and the pre-training details in Tables \ref{tab:pt_hyper1}, \ref{tab:pt_hyper2}.

\begin{table}[tp]
    \centering
    \setlength\tabcolsep{8.0pt}
    \resizebox{1.0\linewidth}{!}{
        \begin{tabular}{c|c|c}
        \textbf{Stage} & \textbf{ViT-B} & \textbf{Output Size} \\
        \Xhline{1.0pt}
        Data & sparse sampling & \violet{3}$\times$\darkGreen{8}$\times$\myblue{224}$\times$\myblue{224} \\
        \hline
        Patch & \darkGreen{1}$\times$\myblue{14}$\times$\myblue{14}, \violet{768} & \multirow{2}{*}{\violet{768}$\times$\darkGreen{8}$\times$\orange{256}} \\
        Embedding & stride \darkGreen{1}$\times$\myblue{14}$\times$\myblue{14} & ~ \\
        \hline
        Position & sine-cosine & \multirow{2}{*}{\violet{768}$\times$\orange{2048}} \\
        Embedding & \violet{768}$\times$\orange{2048} & ~ \\
        \hline
        \multirow{2}{*}{Mask} & semantic mask & \multirow{2}{*}{\violet{768}$\times$\orange{2048$\cdot$(1-$\rho$)}} \\
        ~ & \textit{mask ratio} $=$ $\rho$ & ~ \\
        \hline
        Encoder & $\left[\begin{array}{c}\text{MHSA(\violet{768})}\\[-.1em] \text{MLP(\violet{3072})}\end{array}\right]$$\times$12 & \violet{768}$\times$\orange{2048$\cdot$(1-$\rho$)} \\
        \hline
        Projection & $\left[\begin{array}{c}\text{LN(\violet{768})}\\[-.1em] \text{MLP(\violet{1408})}\end{array}\right]$$\times$$K$ & $K$$\times$\violet{1408}$\times$\orange{2048$\cdot$(1-$\rho$)} \\
        \end{tabular}
    }
    \vspace{-0.3cm}
    \caption{\textbf{Architecture of video encoder.}
    We take ViT-B with 8-frame input as an example.
    ``MHSA'', ``MLP'' and ``LN'' refer to spatiotemporal multi-head self-attention, multi-layer perceptron and layer normalization.
    $K$ means the layer number for unmasked token alignment.
    We mark the \violet{channel number}, \darkGreen{frame number}, \myblue{spatial size} and \orange{token number} by different colors.
    }
    \label{tab:model_architecture}
\end{table}

\begin{algorithm*}[t]
\definecolor{codeblue}{rgb}{0.25,0.5,0.5}
\definecolor{codekw}{rgb}{0.85, 0.18, 0.50}
\begin{lstlisting}[language=python]
class DiffusionLoss(nn.Module)
    def __init__(depth = 6, width = 1536):
        # SimpleMLP takes in x_t, timestep, and condition, and outputs predicted noise.
        self.net = SimpleMLP(depth, width)

        # GaussianDiffusion offers forward and backward functions q_sample and p_sample.
        self.diffusion = GaussianDiffusion()

    # Given condition z and ground truth token x, compute loss
    def loss(self, z, x):
        # sample random noise and timestep
        noise = torch.randn(x.shape)
        timestep = torch.randint(0, self.diffusion.num_timesteps, x.size(0))

        # sample x_t from x
        x_t = self.diffusion.q_sample(x, timestep, noise)

        # predict noise from x_t
        noise_pred = self.net(x_t, timestep, z)

        # L2 loss
        loss = ((noise_pred - noise) ** 2).mean()

        return loss
\end{lstlisting}
\caption{
\textbf{Pseudo-code illustrating the concept of Diffusion Loss}.
Here, the conditioning vector $z$ is the output from the Predictor. The gradient is backpropagated to $z$. The GT $x$ is generated by patching the video input with a pixel shuffle function and a patch size of 14. And then we input the masked part of $x$ and the predicted part $z$ into the model.
}
\label{alg:diffloss}
\end{algorithm*}

\begin{table}[t!]
    \centering
    \setlength\tabcolsep{6pt}
    \resizebox{\linewidth}{!}{
        \begin{tabular}{l|cc}
        config & SthSth V2 & Others \\
        \Xhline{1.0pt}
        optimizer & \multicolumn{2}{c}{AdamW \cite{adamw}} \\ 
        optimizer momentum & \multicolumn{2}{c}{$\beta_1, \beta_2{=}0.9, 0.98$}  \\
        weight decay & \multicolumn{2}{c}{0.05} \\
        learning rate schedule & \multicolumn{2}{c}{cosine decay~\cite{cosine}} \\
        learning rate & \multicolumn{2}{c}{1e-3}\\
        batch size & \multicolumn{2}{c}{2048} \\
        warmup epochs \cite{warmup} & \multicolumn{2}{c}{5} \\
        total epochs &  \multicolumn{2}{c}{50} \\
        input frame & \multicolumn{2}{c}{16} \\
        spatial resolution & \multicolumn{2}{c}{224} \\
        flip augmentation & \textit{no} & \textit{yes} \\
        augmentation & \multicolumn{2}{c}{MultiScaleCrop [0.66, 0.75, 0.875, 1]} \\
        \end{tabular}
    }
    \caption{
        \textbf{InternVideo-Next Stage 1 pre-training settings.}
    }
    \label{tab:pt_hyper1} 
\end{table}

\begin{table}[t!]
    \centering
    \setlength\tabcolsep{6pt}
    \resizebox{\linewidth}{!}{
        \begin{tabular}{l|cc}
        config & SthSth V2 & Others \\
        \Xhline{1.0pt}
        optimizer & \multicolumn{2}{c}{AdamW \cite{adamw}} \\ 
        optimizer momentum & \multicolumn{2}{c}{$\beta_1, \beta_2{=}0.9, 0.98$}  \\
        weight decay & \multicolumn{2}{c}{0.05} \\
        learning rate schedule & \multicolumn{2}{c}{cosine decay~\cite{cosine}} \\
        learning rate & \multicolumn{2}{c}{1e-4}\\
        batch size & \multicolumn{2}{c}{2048(Base) 1024(Large)} \\
        warmup epochs \cite{warmup} & \multicolumn{2}{c}{0} \\
        total epochs &  \multicolumn{2}{c}{100} \\
        input frame & \multicolumn{2}{c}{32} \\
        spatial resolution & \multicolumn{2}{c}{224} \\
        flip augmentation & \textit{no} & \textit{yes} \\
        augmentation & \multicolumn{2}{c}{MultiScaleCrop [0.66, 0.75, 0.875, 1]} \\
        \end{tabular}
    }
    \caption{
        \textbf{InternVideo-Next Stage 2 pre-training settings.}
    }
    \label{tab:pt_hyper2} 
\end{table}

\begin{table}[t!]
    \centering
    \setlength\tabcolsep{2pt}
    \resizebox{0.85\linewidth}{!}{
        \begin{tabular}{l|ccc}
        config & SthSth v2 & Kinetics & Coin \\
        \Xhline{1.0pt}
        optimizer & \multicolumn{3}{c}{AdamW \cite{adamw}} \\ 
        optimizer momentum & \multicolumn{3}{c}{$\beta_1, \beta_2{=}0.9, 0.999$}  \\
        learning rate schedule & \multicolumn{3}{c}{cosine decay~\cite{cosine}} \\
        learning rate & \small{1e-3} & \small{3e-4} & \small{2e-4} \\
        batch size & \multicolumn{3}{c}{1024} \\
        repeated augmentation & \multicolumn{3}{c}{2} \\
        warmup epochs \cite{warmup} & \multicolumn{3}{c}{5} \\
        total epochs &  \small{30} & \small{20} & \small{20} \\
        layer-wise lr decay \cite{beit} & \multicolumn{3}{c}{0.75 (B), 0.85 (L)} \\
        flip augmentation & \textit{no} & \textit{yes} & \textit{yes} \\
        label smoothing \cite{label_smmoth} & \multicolumn{3}{c}{0.1} \\
        cutmix \cite{cutmix} & \multicolumn{3}{c}{1.0} \\
        augmentation & \multicolumn{3}{c}{RandAug(9, 0.5) \cite{randaugment}} \\
        \end{tabular}
    }
    \vspace{-0.3cm}
    \caption{
        \textbf{Action recognition fine-tuning settings.}
    }
    \label{tab:ar_hyperparameters} 
\end{table}

\subsection{Action Recognition}
We show the detailed hyperparameters for action recognition probing in Table \ref{tab:ar_hyperparameters}. We use an attention pooling head that averages the input tokens into a single token. And the single token is utilized as the common [CLS] token for classification, as shown in Figure \ref{fig:supp_main} (a).

\subsection{Depth Estimation}

\paragraph{Experiment Setup.}
To assess the spatiotemporal, low-level, and 3D-geometry related capabilities of InternVideo-Next, we evaluate it on two widely used monocular video depth-estimation benchmarks: ScanNet and KITTI. ScanNet is a large-scale RGB-D video dataset of indoor scenes, containing over 2.5 million views across more than 1500 scans with rich annotations. KITTI is an outdoor dataset collected using an autonomous-driving platform, where stereo videos are recorded by high-resolution color cameras and depth ground truth is captured via a Velodyne LiDAR sensor. Throughout experiments, we adopt a head-probing setup in which the backbone network remains frozen.

\paragraph{Training Details.}
To provide a comprehensive evaluation, we employ two representative prediction heads:  
(1) a \emph{Simple} probing head, following Scaling4d\cite{scaling4drepresentations}, implemented as a lightweight cross-attention pooling layer with an embedding dimension of 1536 (see Figure~\ref{fig:supp_main}(b));  
(2) a \emph{Temporally Aware} head \cite{videodepthanything} based on VideoDepthAnything, consisting of four stacked temporal modules specifically designed to adapt image models such as DINO to video depth-estimation tasks (see Figure~\ref{fig:supp_main}(c)). As it contains spatial modules tailored for depth prediction, we also apply it to video models.  
We use a learning rate of \(3\times 10^{-3}\) with batch sizes of 128 (KITTI) and 256 (ScanNet). Models are trained for 30 epochs, which we find sufficient for convergence.  
A sliding-window strategy is applied to segment input videos, using window sizes of 32 (stride 10) for KITTI and 16 (stride 5) for ScanNet. Following standard practice, the valid depth range is set to \([0.1, 80.0]\) for KITTI and \([0.001, 10.0]\) for ScanNet, with values outside these intervals masked or clamped.

\paragraph{Competitive Methods.}
We compare InternVideo-Next with state-of-the-art image and video representation models, re-implementing all baselines under a unified training and evaluation protocol for fairness.  
For image-based models, we include SigLip2 and DINOv3, known for strong semantic alignment and high-quality visual representations.  
For video-based models, we evaluate VideoMAEv2, InternVideo2, and V-JEPA, representing the current state of the art in video representation learning.

\paragraph{Evaluation Metrics.}
Following standard monocular depth-estimation protocols, we report two metrics that jointly assess geometric accuracy.

\textbf{1. Absolute Relative Error (AbsRel):}
\[
\text{AbsRel} = \frac{1}{N} \sum_{i=1}^{N} \left( \frac{|D_i - \hat{D}_i|}{D_i} \right),
\]
where \(D_i\) and \(\hat{D}_i\) denote the ground-truth and predicted depths, respectively. This metric reflects the average relative deviation of predictions from the ground truth.

\textbf{2. \(\delta_1\):}
The proportion of predictions within a factor of 1.25 of the ground truth, providing a measure of prediction robustness and can be viewed as the accuracy rate.

\subsection{Object Tracking}

\paragraph{Setup.}
To evaluate the model's capability to capture object-level motion, we conduct experiments on the Waymo Open dataset, following the protocol of Scaling4D \cite{scaling4drepresentations}. The videos include 2D and 3D bounding-box annotations. We use only the RGB frames as model input and the 2D bounding boxes for computing losses and evaluation metrics. Because the original Scaling4D benchmark construction code is not publicly available, we reconstruct the training and test sets based on descriptions provided in their paper. Given the target objects' bounding boxes in the first frame, the models are required to output their bounding boxes in the subsequent frames.

\paragraph{Preprocessing.}
Raw \(1280\times 1920\) RGB videos are spatially downsampled to \(224\times 336\) and then centrally cropped to \(224\times 224\). All bounding boxes are remapped to the cropped coordinate space, and boxes occupying less than 0.5\% of the cropped frame area are filtered out. Temporally, the original 20-second sequences recorded at 10~fps are further downsampled to 5~fps.

\paragraph{Dataset Construction.}
Training samples are generated via a sliding window of 16 frames (length \(=16\), stride \(=1\)) applied to each downsampled video. To avoid excessive redundancy, the starting indices for extracted windows must be at least three frames apart. We impose several object-consistency constraints: each window must contain 1--25 objects in the first frame; each first-frame object must appear in at least 70\% of frames within the window; all first-frame objects must co-occur in at least 10 frames; and each object may have at most a continuous 5-frame missing gap. No further area-based filtering beyond the initial 0.5\% threshold is applied.

\paragraph{Training and Evaluation.}
During training, the backbone is frozen and only the probing head is fine-tuned. The head encodes the initial bounding boxes---specifying the objects to track---using a query encoder. The resulting query tokens are fed into a one-layer cross-attention module to obtain pooled tokens, which are then decoded linearly to regress bounding boxes for all frames except the first. The training objective is a weighted sum of Smooth L1 loss (1.0) and GIoU loss (0.5).  
For evaluation, the Intersection-over-Union (IoU) is averaged across all objects and frames.

\section*{Acknowledgement}
This work is supported by the National Key R\&D Program of China (No. 2022ZD0160900), and the Basic Research Program of Jiangsu (No. BK20250009).
{
    \small
    \bibliographystyle{ieeenat_fullname}
    \bibliography{main}
}


\end{document}